\pdfoutput=1
\documentclass[11pt]{article}

\usepackage[final]{acl}
\PassOptionsToPackage{table,dvipsnames}{xcolor}

\usepackage{times}
\usepackage{latexsym}

\usepackage{graphicx}    % 用于插入图片
\usepackage[T1]{fontenc}
\usepackage[utf8]{inputenc}

\usepackage{microtype}

\usepackage{inconsolata}

\usepackage[ruled]{algorithm2e}
\usepackage{multirow}
\usepackage{latexsym}
\usepackage{amsmath,amssymb,amsthm}
\usepackage{subfigure}
\usepackage{microtype}
\DeclareGraphicsExtensions{.pdf, .png}

\usepackage{caption}
\usepackage{subcaption}
\usepackage{array}

\usepackage{booktabs}
\usepackage{setspace}
\usepackage[normalem]{ulem}
\useunder{\uline}{\ul}{}
\usepackage{tikz}
\usepackage{bbding}
\usepackage{pifont}
\usepackage{wasysym}
\usepackage{paralist}
\usepackage{setspace}
\usepackage{adjustbox}

\usepackage{graphicx}
\usepackage{subcaption}
\usepackage{tikz}
\usepackage{pgf-pie}
\usepackage{pgfplots}

\title{PgM\includegraphics[width=1.25em]{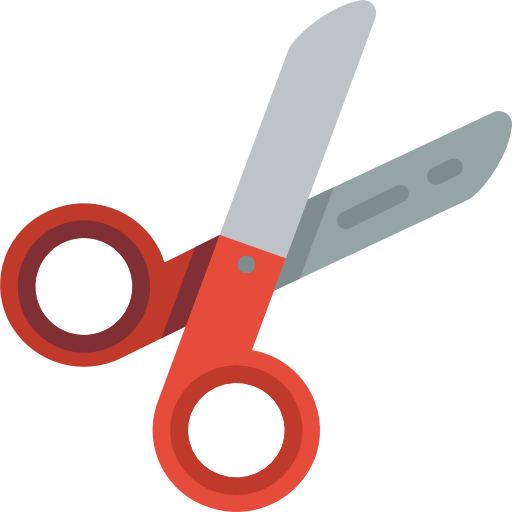}: Partitioner Guided Modal Learning Framework}

\author{Guimin Hu$^{1,2}$, Yi Xin$^{3}$, Lijie Hu$^{4}$, Zhihong Zhu$^{5}$, Hasti Seifi$^{6}$  \\
  $^1$Guangdong University of Technology \\
  $^2$University of Copenhagen \\
  $^3$Nanjing University \\
  $^4$Mohamed bin Zayed University of Artificial Intelligence \\
  $^5$Tencent \\
  $^6$Arizona State University \\
  \texttt{rice.hu.x@gmail.com}}

\begin{document}

\maketitle

%% 主要解决的问题
%% 1.uni-modal feature学习不充分问题, uni-modal feature只关注学习模态内部、自身的学习
%% 2.paired feature学习与模态的交互关系
%% 3.模态学习中uni-modal feaure与paired feature学习过程中的balance关系。
%% 4.将模态表示分解为uni-modal feature和paired feature，更加灵活地适配下游任务。比如，有的下游任务更关注与uni-modal feature学习，而有的下游任务则更关注paired feature学习, 通过将modal representation分解为uni-modal feature和paired feature来更灵活地适应下游任务。
%% 5. 基于下游任务控制每个模态下uni-modal feature和paired feature学习的程度。
%%%贡献点：
%% 1. 提出一种基于分区器的多模态学习方法，其由模态分区器和模态学习器组成。模态分区器将模态表示分为uni-modal feature和paired-modal feature, 模态学习器则分别基于uni-modal和paired-modal feature学习, 其中，uni-modal learner关注模态内与自身相关的表征学习，而paired-modal learner则关注那些与其他模态交互的表征学习。
\begin{abstract}
Multimodal learning benefits from multiple modal information, and each learned modal representations can be divided into \textit{uni-modal} that can be learned from uni-modal training and \textit{paired-modal} features that can be learned from cross-modal interaction. Building on this perspective, we propose a partitioner-guided modal learning framework, PgM\includegraphics[width=1.25em]{figures/cut_2.png}, which consists of the modal partitioner, uni-modal learner, paired-modal learner, and uni-paired modal decoder. Modal partitioner segments the learned modal representation into uni-modal and paired-modal features. Modal learner incorporates two dedicated components for uni-modal and paired-modal learning. Uni-paired modal decoder reconstructs modal representation based on uni-modal and paired-modal features. {\bf PgM offers three key benefits}: 1) thorough learning of uni-modal and paired-modal features, 2) flexible distribution adjustment for uni-modal and paired-modal representations to suit diverse downstream tasks, and 3) different learning rates across modalities and partitions. Extensive experiments demonstrate the effectiveness of PgM across {\bf four multimodal tasks} and further highlight its transferability to existing models. Additionally, we visualize the distribution of uni-modal and paired-modal features across modalities and tasks, offering insights into their respective contributions.
\end{abstract}


%%%%%%%%%%%%%%%%%%%%%%%%%%%%%%%%%%%%%%%%%%%%%%%%%%%%%%%%%%%%%%%%%%%%
\section{Introduction}

Multimodal data is widely used to enhance machine learning systems, which integrates these diverse modalities to improve decision-making accuracy \cite{hu2022unimse,zhu2024towards,zhu2024tfcd, hu2024unimeec,xie2025explicitly,li2024foodieqa}. With the inclusion of additional information, multimodal networks are expected to match or surpass their uni-modal counterparts, as they leverage richer information from multiple input modalities \cite{DBLP:conf/nips/HuangDXCZH21}. 

However, multimodal networks are often observed to underperform uni-modal networks. This observation is consistent across various modality combinations, tasks, and benchmarks \cite{DBLP:conf/nips/HuangDXCZH21,DBLP:conf/cvpr/WangTF20,DBLP:conf/icml/HuangLZYH22}. \citet{DBLP:journals/corr/abs-2207-00056} analyzes multimodal model behavior by studying the uni-modal importance, cross-modal interactions and so on, and proposes that the gradients of different modalities should be further adjusted during training. \citet{DBLP:conf/icml/DuTLLYWYZ23} also notice the phenomenon of modality laziness, which causes insufficient modal learning of uni-modal feature. \citet{DBLP:conf/cvpr/WangTF20} identifies multimodal network are ofen prone to overfiting due to their increased capacity and different modalities overfit and generalize at different rates. 
\begin{figure}[t]
\centering
\includegraphics[width=0.9\linewidth]{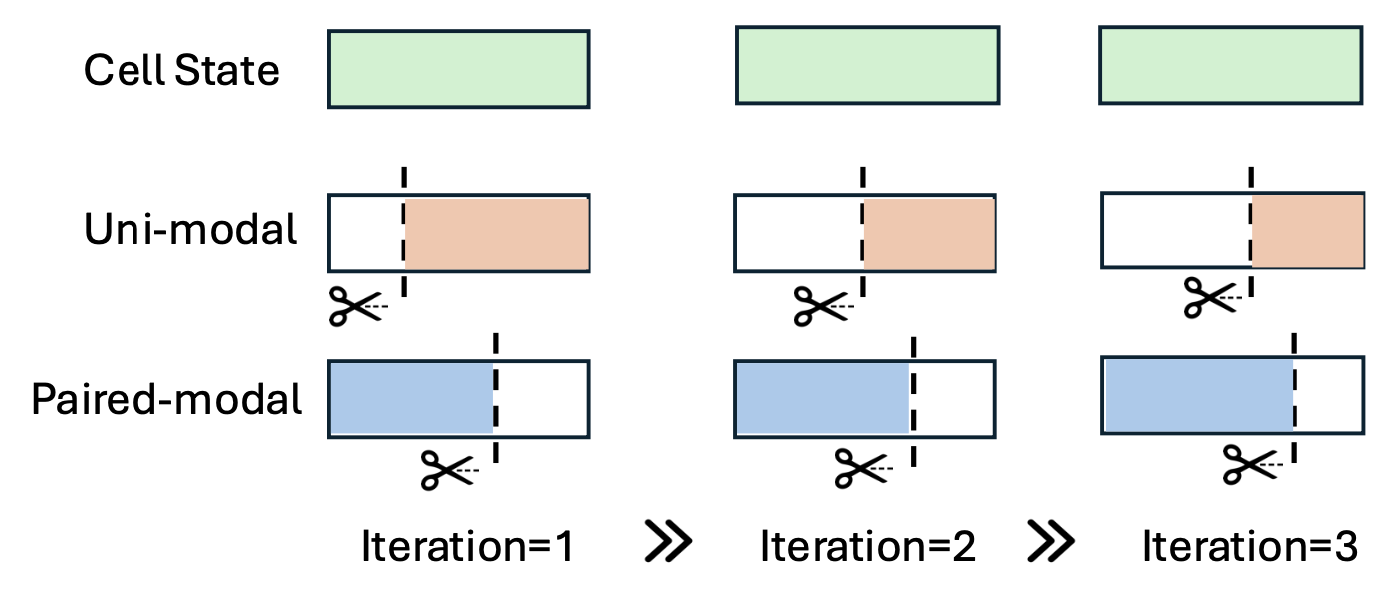}
\caption{Illustration of modal partitioner. Modal partitioner adjusts the distribution of uni-modal and paired-modal representations across multiple iterations.}
\label{fig:example}
\end{figure}

Previous multimodal learning studies summarize three major unresolved challenges. {\bf First}, different types of features in multimodal data are learned at varying rates. Furthermore, The uni-modal and paired-modal representations required for each modality vary across different tasks \cite{DBLP:conf/cvpr/WangTF20,DBLP:conf/icml/DuTLLYWYZ23}. {\bf Second}, achieving an effective multimodal fusion representation requires both robust learning of uni-modal and paired-modal features and dynamic adjustment of their distributions throughout the representation learning process \cite{DBLP:conf/nips/HuangDXCZH21}. {\bf Third}, modality laziness causes insufficient modal learning since multimodal networks are often prone to overfitting and generalizing at different rates \cite{DBLP:conf/icml/DuTLLYWYZ23}. 

Considering how features (i.e., learned representations) from multimodal data are acquired in supervised learning, we propose a modal partitioner, as illustrated in Figure \ref{fig:example}, where uni-modal and paired-modal partitions may overlap and are not mutually exclusive. From this perspective, we propose a {\bf P}artitioner-{\bf G}uided {\bf M}odal learning framework, {\bf PgM}\includegraphics[width=1.25em]{figures/cut_2.png}, which comprises a modal partitioner for segmentation, two modal learner for modal partition learning, and a modal decoder for modality reconstruction. The modal partitioner first segments the learned modal representation into uni-modal and paired-modal partitions. Next, the modal learner includes two dedicated components for learning uni-modal and paired-modal representations, while the modal decoder reconstructs modality information between these features to effectively capture their characteristics. Finally, PgM is integrated with downstream tasks in an end-to-end architecture, enabling joint training. {\bf PgM offers three key advantages}: 1) it enables efficient learning of both uni-modal and paired-modal features, 2) it allows for flexible adaptation of feature distributions across different downstream tasks while dynamically adjusting uni-modal and paired-modal contributions, and 3) it tailors learning rates across modalities and partitions to counteract modality laziness. To sum up, our contributions are three-folds:
\begin{compactitem}
    \item[1.] We propose PgM\includegraphics[width=1.25em]{figures/cut_2.png}, a partitioner-guided modal learning framework, consists of the modal partitioner, two modal learners (e.g., uni-modal and paired-modal learners), and a uni-paired modal decoder for effective modal partition representation learning. 
    \item[2.] PgM can be integrated into existing models for downstream multimodal tasks, demonstrating its transferability and feasibility. 
    \item[3.] Experimental results demonstrate the effectiveness of PgM across four multimodal tasks and its transferability to existing models, providing deeper insights into multimodal learning.
\end{compactitem}

\section{Related Work}
\subsection{Multimodal Networks}
Multimodal networks process multimodal signals as input and integrate the information to support decision-making in downstream tasks. These networks either use one modality as input to predict another (e.g., Visual-Q\&A \cite{DBLP:conf/cvpr/GoyalKSBP17}) or leverage cross-modality correspondences for self-supervised learning (e.g., image-audio correspondence \cite{DBLP:conf/iccv/ArandjelovicZ17}). Existing multimodal networks can be broadly classified into three categories: Transformer-based architectures, disentangle-based methods and generation-based structures. First is Transformer-based architectures. For instance, \citet{DBLP:conf/acl/TsaiBLKMS19} introduces the Multimodal Transformer (MulT), which addresses multimodal fusion in an end-to-end manner. Second is disentangle-based methods. These approaches separate modal representations into modality-invariant and modality-specific feature spaces. For example, \citet{DBLP:conf/aaai/ZhangCSW22} introduces an adversarial multimodal refinement module to explore shared characteristics across modalities while enhancing each modality's uniqueness. Third is generation-based structures that utilize translation modules to generate modal representations from another modality. More recently, models like BLIP \cite{li2022blip} and BLIP-2 \cite{li2023blip} have been proposed for vision-language tasks, excelling in both understanding and generation.

\subsection{Multimodal Learning}
Multimodal learning involves integrating information from different modalities, which can be categorized into three types: early fusion, late fusion, and hybrid fusion. Early fusion aggregates the raw features from different modalities into a joint representation before modal encoder \cite{perez2019mfas}. Late fusion combines the decisions from different classifiers into one final decision \citet{10023506}. Compared with early fusion and late fusion, hybrid fusion is a multimodal learning approach that combines elements of both early fusion and late fusion, aiming to leverage the strengths of both methods while mitigating their weaknesses \cite{hu2022unimse,zhu2024towards,hu2024unimeec}.
Recently, \citet{DBLP:conf/cvpr/WangTF20} observes that the best-performing uni-modal network often outperforms its multimodal counterpart. This phenomenon is consistent across various modality combinations, tasks, and benchmarks in video classification. Similarly, \citet{DBLP:conf/icml/DuTLLYWYZ23} highlights the issue of modality laziness, where modal representation fails to learn effectively. \citet{DBLP:conf/icml/HuangLZYH22} further demonstrates that in multimodal late-fusion networks with (smoothed) ReLU activation trained via gradient descent, different modalities compete, leading the encoder networks to focus on only a subset of modalities.

The prior work underscore that different modalities exhibit varying learning dynamics during training, influencing final model performance. Some studies \cite{DBLP:journals/corr/abs-2207-00056} analyze multimodal model behavior by examining uni-modal importance, cross-modal interactions, and other factors, arguing that modality-specific gradient adjustments are necessary. Additionally, uni-modal features in multimodal fusion representations may be insufficient due to significant redundancy—often repetitive or uninformative—in the final fused representation \cite{DBLP:conf/nips/HuangDXCZH21}.

%%%%%%%%%%%%%%%%%%%%%%%%%%%%%%%%%%%%%%%%%%%%%%%%%%%%%%%%%%%%%%%%%%%%%
\section{Methodology}
\begin{figure*}[t]
\centering
\subfigure[]{\includegraphics[width=0.33\textwidth]{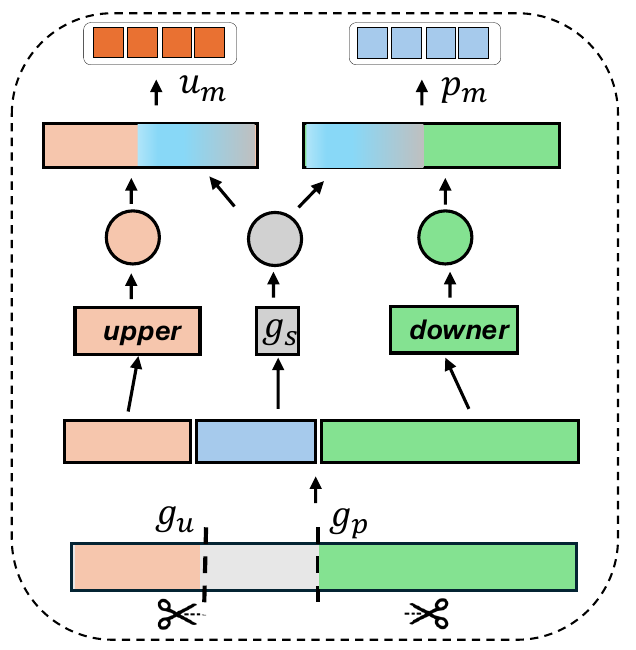}\label{fig:segmentation}}
\subfigure[]{\includegraphics[width=0.58\textwidth]{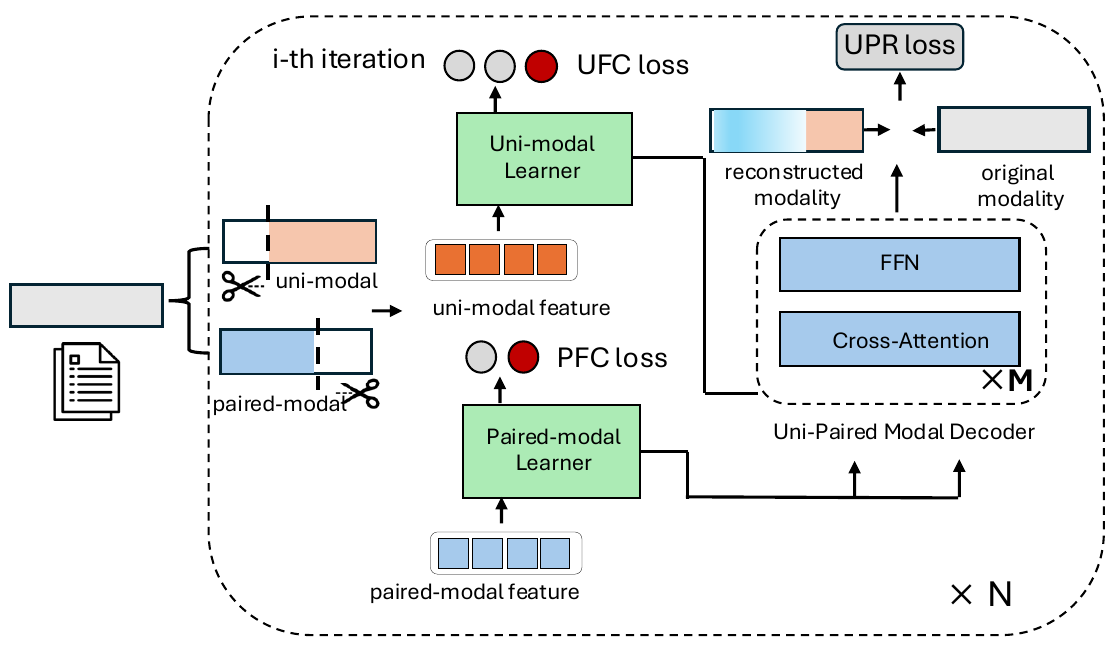}\label{fig:architecture}}
\hspace{0.05\textwidth}
\caption{Overview of PgM: (a) Segmentation process of partitioner,(b)Architecture of PgM.}
\label{fig:overview}
\end{figure*}

\subsection{Modal Partitioner}
Given a sample containing multiple modalities $\{m^{1}, m^{2}, \cdots, m^{N}\}$ (e.g., text modalities), we set $N$ individual modal encoders to map raw modal signal into a $d$-dimensional representation vector. After modal encoder, we derive the modality representation set $\{\mathbf{I}_{m^{1}}, \cdots, \mathbf{I}_{m^{N}}\}$ for $N$ modalities. Figure \ref{fig:overview}(a) provides segmentation process by the partitioner.

The modal partitioner processes modal representations by partitioning neurons into two groups: uni-modal and paired-modal features. The uni-modal features store information for uni-modal training, while the paired-modal features capture information for cross-modal interactions \cite{DBLP:conf/icml/DuTLLYWYZ23}. We assign a modal partitioner to each modality, dividing each modal representation into two partitioners. This allows adaptive adjustment of uni-modal and paired-modal feature distributions based on the model's downstream task performance. Given a modality representation $\mathbf{I}_m$, the segmentation point between uni-modal and paired-modal features is determined using the cumulative softmax activation function, defined as $\text{cumsoftmax}(\cdot)=\text{cumsum}(\text{softmax}(\cdot))$\footnote{$\text{cumsoftmax}(x_i) = \sum_{j=1}^{i} \frac{e^{x_j}}{\sum_{k=1}^{n} e^{x_k}}$s}. This function approximates a binary gating mechanism in the form of $(0,\cdots, 1,\cdots1)$. In the $i$-th iteration, partitioner gates for uni-modal and paired-modal are defined as follows:
\begin{align}
\begin{split}
    &\mathbf{g}_u^{(i)} = 1 - \text{cumsoftmax}(\mathbf{u}_m)\\
    &\mathbf{g}_p^{(i)} = \text{cumsoftmax}(\mathbf{p}_m)
\end{split}
\end{align}
Initially, in the first iteration, $\mathbf{u}_{m}=\mathbf{p}_{m}=\mathbf{I}_{m}$, $m\in \{m^{1},m^{2},\cdots,m^{N}\}$ denotes specific modality. 
$\mathbf{g}_u^{(1)}$ and $\mathbf{g}_p^{(1)}$ denote the uni-modal and paired-modal gates in the first iteration, respectively. In the $i$-th iteration,$\mathbf{g}_u^{(i)}\in \mathcal{R}^{1\times D}$ follows the form $[u_{1},\cdots,u_{r},0,\cdot,0]$, and $\mathbf{g}_p^{(i)}\in \mathcal{R}^{1\times D}$ takes the form $[0,\cdot,0,p_{1},\cdots,p_{l}]$. These gates control the uni-modal and paired-modal features in the $i$-th iteration, where $u_{i}\to1$ and $p_{i}\to1$. 
\begin{align}
\begin{split}
    &\mathbf{g}_s^{(i)} = \mathbf{g}_u^{(i)} \circ \mathbf{g}_p^{(i)}\\
    &\mathbf{upper}_{m}^{(i)} = \mathbf{g}_u^{(i)} - \mathbf{g}_s^{(i)}\\
    &\mathbf{downer}_{m}^{(i)} = \mathbf{g}_p^{(i)} - \mathbf{g}_s^{(i)}
\end{split}
\end{align}
where $\circ$ denotes element-wise multiplication, $\mathbf{g}_{s}^{(i)}$ represents the gate overlap between uni-modal and paired-modal features. $\mathbf{upper}_{m}^{(i)}$ refers to the upper part of the cell neurons representing uni-modal features, while $\mathbf{downer}_{m}^{(i)}$ refers to the lower part of the cell neurons representing paired-modal features. $\mathbf{I}^{m}_{s}$ denotes the shared modal representation between uni-modal and paired-modal. Subsequently, we aggregate partition information from both target cells to obtain the updated uni-modal and paired-modal representations:
\begin{align}
    \begin{split}
    &\mathbf{I}^{m}_{s} = \mathbf{g}_s^{(i)} \circ \mathbf{I}_{m}\\
    &\mathbf{u}_{m}^{(i+1)} = \mathbf{upper}_c \circ \mathbf{u}_{m}^{(i)} + \mathbf{I}^{m}_{s}\\
    &\mathbf{p}_{m}^{(i+1)} = \mathbf{downer}_c \circ \mathbf{p}_m^{(i)} + \mathbf{I}^{m}_{s}
    \end{split}
\end{align}
where $\{\mathbf{u}_{m}^{(i+1)}, \mathbf{p}_{m}^{(i+1)}\}$ are updated uni-modal and paired-modal features after the $i$-th iteration, respectively. Thus, after $N$ iterations, the modal partitioner divides each learned modal representation into the uni-modal partition $\mathbf{u}_{m}^{(N)}$ and the paired-modal partition $\mathbf{p}_{m}^{(N)}$.

\subsection{Modal Learner and Decoder}
The modal learner consists of (1) uni-modal learner and (2) paired-modal learner, and each learner contains multiple stacked Transformer blocks. To enhance uni-modal and paired-modal learning, we apply padding masks to specific neurons within the multimodal representation, ensuring that the uni-modal learner and paired-modal learner focus exclusively on uni-modal and paired-modal features, respectively. In the $i$-th iteration, the padding masks are derived from $\mathbf{g}_{u}^{(i)}$ and $\mathbf{g}_{p}^{(i)}$ to tailor the attention masks for the two learners. For example, $\mathbf{g}_{u}^{(i)}=[u_{1},\cdots,u_{m},0,\cdots,0]$, $\mathbf{g}_{p}^{(i)}=[0,\cdot,0,p_{1},\cdots,p_{m}]$, and $u_i\to 1$, $p_i\to 1$, thus we can approximately formalize $\mathbf{g}_{u}^{(i)}$ and $\mathbf{g}_{p}^{(i)}$ as $[1,\cdots,1,0,\cdots,0]$ and $[0,\cdots,0,1,\cdots,1]$, denoted by $\hat{\mathbf{g}}^{(i)}_{u}$ and $\hat{\mathbf{g}}^{(i)}_{p}$, respectively:
% to $[1,\cdots,1,0,\cdots,0]$ and converting $\mathbf{g}_{p}^{(i)}$ as $[0,\cdots,0,1,\cdots,1]$.
\begin{align}
    &\mathbf{M}_{z}^{(i)} = (\mathbf{1.0} - \hat{\mathbf{g}}^{(i)}_{z}) \cdot C,\quad z\in \{u, p\}\\
    &\mathbf{A}_{z}^{(i)} = \text{softmax}(\frac{\mathbf{Q}\mathbf{K}^{T}}{\sqrt{d}}+ \mathbf{M}_{z}^{(i)})
\end{align}
where constant $C=-10000.0$, $\mathbf{M}_{z}^{(i)}$ denotes padding mask, $\{\mathbf{A}_{u}^{(i)}, \mathbf{A}_{p}^{(i)}\}$ represents the attention weight matrix applied to the Transformer \cite{vaswani2017attention} modules for the uni-modal and paired-modal learners, respectively. 
% Both learners take Transformer as backbone to implement effectively uni-modal and paired-modal representation training.

\paragraph{Uni-modal and Paired-modal Learners} uni-modal learner focuses on intra-uni-modal feature learning, and paired-modal learner attends on intra-paired-modal learning. Both learners take Transformer as backbone, which takes modal representation as query, key and value. Noted that uni-modal and paired-modal learners takes $\mathbf{M}^{u}\in \mathcal{R}^{S\times D}$ and $\mathbf{M}^{p}\in \mathcal{R}^{S\times D}$ as padding masks respectively, ensuring the learners only focus on uni-modal and paired-modal features. Here, $S$ and $D$ denote the sequence length and representation dimension, respectively.
% \begin{align*}
% \mathbf{M}^{u} =\begin{bmatrix}
% 0 & \cdots& 0 &  -\infty & \cdots & -\infty \\
% 0 & \cdots& 0 &  -\infty & \cdots & -\infty \\
% 0 & \cdots& 0 &  -\infty & \cdots & -\infty \\
% 0 & \cdots& 0 &  -\infty & \cdots & -\infty \\
% \end{bmatrix}\\
% \\
% \mathbf{M}^{p} =\begin{bmatrix}
%  -\infty & \cdots & -\infty & 0 & \cdots& 0\\
% -\infty & \cdots & -\infty & 0 & \cdots& 0\\
% -\infty & \cdots & -\infty & 0 & \cdots& 0\\
% -\infty & \cdots & -\infty & 0 & \cdots& 0\\
% \end{bmatrix}
% \end{align*}
\begin{figure}[h]
\vspace{-2pt}
\centering
\includegraphics[width=0.83\linewidth]{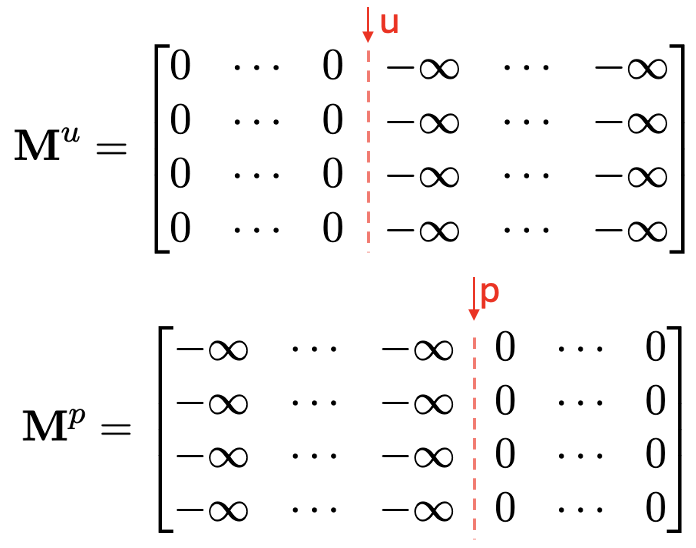}
\end{figure}
The padding masks for uni-modal features (i.e., $\mathbf{M}^{u}$) and paired-modal features (i.e., $\mathbf{M}^{p}$) are defined as follows, respectively:
\begin{align}
  &{M}_{*j}^{u} =
\begin{cases}
0, & j <= u \\
-\infty, &  j > u
\end{cases}\\
&M_{*j}^{p} =
\begin{cases}
0, & j >= p\\
-\infty, & j < p
\end{cases}  
\end{align}
where $u$ and $p$ denote the upper and lower segmentation points of the uni-modal and paired-modal partitions, respectively. Both are derived from $\hat{\mathbf{g}}^{(i)}_{u}$ and $\hat{\mathbf{g}}^{(i)}_{p}$ (see Appendix \ref{sec:architecture_pgm} for details).

% Based on uni-modal and paired-modal learners, we then establish an uni-paired modal decoder for original modal representation reconstruction. 

\paragraph{Uni-paired Modal Decoder} The uni-paired modal decoder takes the concatenated uni-modal and paired-modal features to reconstruct the modal representation.
\begin{align}
\hat{\mathbf{I}}_{m}^{(i)} = \text{Decoder}([\mathbf{u}_{m}^{(i)},\mathbf{p}_{m}^{(i)}])
\end{align}
where $[,]$ denotes concatenation operation, $\hat{\mathbf{I}}_{m}^{(i)}$ denotes the reconstructed modal representation produced by uni-paired modal decoder.

\subsection{Pre-training Objectives}
The PgM architecture is illustrated in Figure \ref{fig:overview}(b). The framework is trained with three objectives: {\bf Uni-modal Feature Classification}, which enhances uni-modal feature learning; {\bf Paired-modal Feature Classification}, which focuses on paired-modal feature learning; and {\bf Uni-paired Modal Reconstruction}, which reconstructs modalities by leveraging both uni-modal and paired-modal features within the same modality.

\paragraph{Uni-modal Feature Classification} aims to learn better uni-modal representations before performing modal fusion. We feed uni-modal feature of each iteration into classifier to estimate which modality the representation comes from, which ensures the uni-modal features store the essential information for discriminability of modality. For a sample with two modalities (e.g., text and vision), the ground truth modality labels of text and vision modalities are denoted as $O^{t}$ and $O^{v}$, respectively:
\begin{align}
    \mathbf{O}^{t}= \begin{bmatrix}
1 & 0 \\ 
\cdot & \cdot \\ 
1 & 0
\end{bmatrix}, 
    \mathbf{O}^{v}= \begin{bmatrix}
0 & 1 \\ 
\cdot & \cdot \\ 
0 & 1
\end{bmatrix}
\end{align}
Similarly, we train an \( N \)-class classifier to determine the modality source of the uni-modal feature for a sample with \( N \) modalities. We denote the uni-modal feature classification loss $\mathcal{L}^{UFC}$ as the main supervision of uni-modal learner.

\paragraph{Paired-modal Feature Classification}
To ensure the paired-modal feature remains distinct from the uni-modal feature, we input the paired-modal feature of each iteration into a classifier designed to differentiate between uni-modal and paired-modal representations. We formalize this process as a binary classification task, where the ground truth modality label is defined as: 
\begin{align}
    \mathbf{O}^{u}= \begin{bmatrix}
1 & 0 \\ 
\cdot & \cdot \\ 
1 & 0
\end{bmatrix}, 
    \mathbf{O}^{p}= \begin{bmatrix}
0 & 1 \\ 
\cdot & \cdot \\ 
0 & 1
\end{bmatrix}
\end{align}
where $\mathbf{O}^{u}$ and $\mathbf{O}^{p}$ denote the ground labels of uni-modal and paired-modal representation, respectively. We denote the paired-modal feature classification loss $\mathcal{L}^{PFC}$ as the main supervision of paired-modal learner.

\paragraph{Uni-Paired Modal Reconstruction}
Each modal representation is decomposed into uni-modal and paired-modal components. The uni-paired modal decoder applies a reconstruction loss to ensure that the reconstructed modal representation captures both uni-modal and paired-modal features simultaneously, effectively reconstructing the original modal representation.
\begin{align}
    \mathcal{L}^{UPR} = \frac{1}{|S|}\left( \sum_{m \in \{\mathbf{{m^{1}}}, \cdots, \mathbf{{m^{N}}}\}} \frac{\|\mathbf{I}_m - \hat{\mathbf{I}}_m\|_2^2}{d_h} \right)
\end{align}
where $|S|$ represents the number of samples in the training set. We define the uni-paired modal reconstruction loss $\mathcal{L}^{UPR}$ as the primary supervisory signal for the uni-paired decoder.

\subsection{Downstream Multimodal Tasks}
During training, we integrate multimodal supervised downstream tasks and the partitioner-guided modal learning framework (PgM) for joint training and evaluate the performance on multimodal tasks. The architecture and additional details are provided in the Appendix \ref{sec:architecture_pgm} and \ref{sec:downstream_training}. The training process comprises two stages: (1) pre-training the partitioner-guided modal learning framework (PgM) with the pre-training objective $\mathcal{L}^{P}$, and (2) fine-tuning PgM using both the downstream task loss and the pre-training objective loss $\mathcal{L}^{D}$:
\begin{align}
&\mathcal{L}^{P}=\sum_{i}^{N} (\mathcal{L}^{UFC}+\mathcal{L}^{PFC}+\mathcal{L}^{UPR})\label{eq:(13)}\\
&\mathcal{L}^{D}=\alpha \mathcal{L}^{P}+\beta \mathcal{L}^{T}\label{eq:(14)}
\end{align}
where $\mathcal{L}^{T}$ represents the training loss for downstream task $T$, $\{\alpha, \beta\}$ are the hyperparameters of PgM, $i$ represents the $i$-th iteration and $N$ is the iteration number. Modal partitioner adjusts the distribution of uni-modal and paired-modal representations across multiple iterations. 

After applying PgM, we obtain uni-modal and paired-modal features for each modality, enabling flexible utilization of these partitioned features. In this paper, we separately concatenate all uni-modal and paired-modal features from different modalities within the same sample and model cross-modal interactions using a Transformer layer. For downstream tasks, we concatenate the output of Transformer as input to the FFN layer for final prediction. The training process consists of two phases: first, we train PgM using the loss function in Equation (\ref{eq:(13)}) for $\text{N}^{1st}$ epochs, then we jointly train PgM and the downstream task using the loss function in Equation (\ref{eq:(14)}) for $\text{N}^{2st}$ epochs. The loss curves for joint training are provided in Appendix \ref{sec:loss_curve}, and module sizes are detailed in Appendix \ref{sec:model_size}.

\begin{table*}[t]
\centering
\resizebox{\textwidth}{!}{\begin{tabular}{llccc}
\toprule
\textbf{\# Task} & \textbf{\# Dataset} & \textbf{\# Total Instances} & \textbf{\# Number of Annotations} & \textbf{\# Modality} \\
\midrule
\multirow{1}{*}{Multimodal Sentiment Analysis} & MOSI  & 2199 & 3 & A+V+T \\
\multirow{1}{*}{Multimodal Emotion Recognition} & MELD  & 9989 & 6 & A+V+T \\
Cross-modal Retrieval & Wikipedia & 2866 & 10 & V+T \\
Image-text Classification & UMPC Food 101 & 90686 & 101 & V+T \\
\bottomrule
\end{tabular}}
\caption{Information about the datasets used in four tasks. \textbf{A}, \textbf{V}, \textbf{T} denote \textbf{Audio}, \textbf{Vision} and \textbf{Text} modality.}
\label{tab:datasets}
\end{table*}
\section{Experiments}
\subsection{Tasks and Datasets}
In this work, we focus on the following tasks:
{\bf Multimodal Sentiment Analysis (MSA)} \cite{hu2022unimse,mao2022m} aims to predict the sentiment polarity by leveraging three types of modalities: audio, vision and text. {\bf Multimodal Emotion Recognition in Conversation (MERC)} aims to predict predefined emotion categories (e.g., joy and sadness) using text, audio, and visual modalities. {\bf Cross-modal Retrieval (CR)} \cite{li2021align,singh2022flava} is the process of finding the relevant items in one modality based on the query in another modality. {\bf Image-text Classification (ITC)} \cite{hu2024unimeec,lian2024mer} involves using both visual and textual information to classify the given information into 101 categories. We evaluate our proposed model on MSA, MERC, CR and ITC using \uline{MOSI} \cite{DBLP:journals/expert/ZadehZPM16}, \uline{MELD} \cite{DBLP:conf/acl/PoriaHMNCM19}, \uline{Wikipedia} \cite{wang2015recipe} and \uline{UMPC Food 101} \cite{rasiwasia2010new} datasets, respectively. The details of the tasks and datasets are shown in Table \ref{tab:datasets}.
\begin{table*}[t]
\centering
\resizebox{\textwidth}{!}{
\begin{tabular}{lcccccccc}
\toprule
\toprule
\multirow{2}{*}{\#Model} & \multicolumn{2}{c}{\#Multimodal Sentiment Analysis} 
& \multicolumn{2}{c}{\#Multimodal Emotion Recognition} 
& \multicolumn{2}{c}{\#Cross-modal Retrieval} 
& \multicolumn{2}{c}{\#Image-text Classification} \\ 
\cmidrule(lr){2-3} \cmidrule(lr){4-5} \cmidrule(lr){6-7} \cmidrule(lr){8-9}
& \multicolumn{2}{c}{MOSI} & \multicolumn{2}{c}{MELD} & \multicolumn{2}{c}{Wikipedia} & \multicolumn{2}{c}{UMPC Food 101}\\ \cmidrule(lr){2-3} \cmidrule(lr){4-5} \cmidrule(lr){6-7} \cmidrule(lr){8-9}
& \bf ACC-2 & \bf F1 & \bf ACC & \bf Weighted F1 & \bf MAP & \bf Precision@10 & \bf ACC & \bf Weighted F1 \\ 
\midrule
Single-Modal (Audio)      & 48.58/50.36 & 44.31/46.47 & 51.44 & 52.72 & - & - & - & - \\
Single-Modal (Visual)     & 65.86/69.72 & 65.97/69.94 & 51.75 & 54.06 & - & - & 71.58 & 72.04 \\
Single-Modal (Text)       & 64.33/66.16 & 63.20/65.32 & 56.21 & 56.38 & - & - & 69.92 & 69.89\\ \hline
Concatenation             & 69.15/70.42 & 67.26/69.17 & 58.64 & 58.16 & 62.84 & 56.29 & 82.43 & 80.49\\
Add                       & 67.61/67.61 & 66.13/66.13 & 53.58 & 54.67 & 61.98 & 55.05 & 81.67 & 81.66 \\
Element-wise Maximum      & 68.16/69.50 & 65.16/66.87 & 51.59 & 52.86 & 61.99 & 56.96 & 80.11 & 80.09\\
Linear-Fusion             & 66.52/67.81 & 64.58/65.36 & 53.72 & 54.89 & 61.54 & 54.14 & 77.10 & 77.05\\
MLP                       & 66.30/67.82 & 65.41/66.57 & 54.16 & 55.88 & 61.71 & 55.06 & 84.82 & 84.85\\ \hline
\textbf{PgM}             & \textbf{84.69/85.39} & \textbf{84.65/85.95} & \textbf{66.69} & \textbf{66.95} & \textbf{73.35} & \textbf{70.82} & \textbf{90.36} & \textbf{91.04} \\ 
\bottomrule
\bottomrule
\end{tabular}}
\caption{Comparison between modal learning baselines and PgM.}
\label{tab:main_result}
\end{table*}
\subsection{Evaluation Metrics}
Following the previous work \cite{hu2022unimse,DBLP:conf/emnlp/HanCP21}, we adopt accuracy (ACC) and weighted F1 score (Weighted F1) to evaluate the performance on MELD and UPMC Food 101 datasets, and we adopt accuracy (ACC-2) and F1 as the metric of PgM on MOSI dataset. For Wikipedia dataset, we use the mean average precision (MAP) and Precision@10 (the average precision of the first 10 retrieved items) as the evaluation metric of cross-modal retrieval task. Due to space limit, experimental setting is presented in Appendix \ref{sec:experimental_setting}.

\subsection{Experimental Setting}
\label{sec:experimental_setting}
We use pre-trained T5-Base \footnote{https://github.com/huggingface/transformers/tree/main\\/src/transformers/models/t5.} as text encoder, ViT\footnote{https://huggingface.co/openai/clip-vit-base-patch32} as visual encoder, AST\footnote{https://huggingface.co/docs/transformers/model\_doc/audio-spectrogram-transformer} as audio encoder. 
The batch size is set to 96, with an overall learning rate of 3e-4. The learning rates for the uni-modal and paired-modal learners are set to 0.0001, while the learning rate for the uni-paired decoder is set to 0.001. In Equation (\ref{eq:(14)}), the hyperparameters are defined as $\alpha = 0.5$, $\beta = 1$ and the iteration number $N=3$. The pre-training epoch for PgM is set to 20 (i.e., $\text{N}^{1st}$=20), and the jointly training epoch for downstream task and PgM is set to 50 (i.e.,$\text{N}^{2st}$=50). We adopt the Transformer architecture as the core backbone for uni-modal and paired modal learners, guided by the partitioner. For decoder, we adapt Cross-Attention layer for reconstruction. For the multimodal task involving acoustic, visual, and textual modalities, the dimensions of learned modal representation for each modality are set to 768, aligning with the fused representation dimension of 768.

\subsection{Baselines}
We set basic modal fusion methods, including Concatenation, Add, Element-wise Maximum, Linear Fusion, MLP, and Uni-Modal Training, as baselines, with the following details: {\bf Concatenation} concatenates feature vectors from different modalities into one feature vector. {\bf Add} adds feature vectors from different modalities into one feature vector. {\bf Element-wise Maximum} selects the element-wise maximum feature among feature vectors from different modalities. {\bf Linear Fusion} applies linear combination of features in feature vectors from different modalities. {\bf MLP} applies multilayer perceptron layer to fuse features from different modalities. {\bf Uni-Modal Training} applies single modality for downstream task learning. Additionally, we further apply the proposed model fusion to the prior models of MSA and MERC tasks. The details of the prior models are as follows: {\bf Self-MM} \cite{DBLP:conf/aaai/YuXYW21} leverages uni-modal representations through multi-task learning to address the multimodal sentiment analysis task. {\bf MMIM} \cite{DBLP:conf/emnlp/HanCP21} hierarchically maximizes mutual information to tackle the multimodal sentiment analysis task. {\bf UniMSE} \cite{hu2022unimse} introduces a unified sharing framework that bridges multimodal sentiment analysis and multimodal emotion recognition to enhance model performance. {\bf UniMEEC} \cite{hu2024unimeec} explores the complementary influence of emotion causes on multimodal emotion recognition.

\begin{table*}[ht]
\centering
\begin{adjustbox}{max width=\textwidth}
\begin{tabular}{lcccccccccc}
\toprule
\multirow{3}{*}{\#Model} & \multicolumn{2}{c}{\#Multimodal Sentiment Analysis} 
&\multicolumn{2}{c}{\#Multimodal Emotion Recognition} 
& \multicolumn{2}{c}{\#Cross-modal Retrieval} & \multicolumn{2}{c}{\#Image-text Classification}\\ \cmidrule(lr){2-3} \cmidrule(lr){4-5}\cmidrule(lr){6-7} \cmidrule(lr){8-9}
& \multicolumn{2}{c}{MOSI} & \multicolumn{2}{c}{MELD} & \multicolumn{2}{c}{Wikipedia} & \multicolumn{2}{c}{UMPC Food 101}\\ \cmidrule(lr){2-3} \cmidrule(lr){4-5} \cmidrule(lr){6-7} \cmidrule(lr){8-9}
                       &\bf ACC-2 &\bf F1 &\bf ACC&\bf Weighted F1 &\bf MAP  &\bf Precison@10 &\bf ACC &\bf Weighted F1\\ \midrule
\textbf{PgM}               & \textbf{84.69/85.39} & \textbf{84.65/85.95} & \textbf{66.69} & \textbf{66.95} & \textbf{73.35} & \textbf{70.82} & \textbf{90.36} & \textbf{91.04} \\
\midrule
-w/o Modal Partitioner    & 48.58/50.69 & 44.31/47.46 & 51.44 & 52.72 & 60.90 &59.18 &59.69 &53.14\\
-w/o Uni-Modal Learner ($\mathcal{L}^{UFC}$)   & 65.86/67.18 & 65.97/67.06 & 51.75& 54.06& 63.23 &63.64 & 70.12 & 70.65\\ 
-w/o Paired-Modal Learner ($\mathcal{L}^{UPC}$)   & 68.14/70.05 & 66.31/68.69 & 54.46& 56.16& 64.68 &65.25 & 73.65 & 72.96 \\ 
-w/o Uni-Paired Decoder ($\mathcal{L}^{UPR}$)   & 69.90/70.69 & 66.37/67.44 & 56.81& 57.32& 64.21 &64.64 & 74.89 & 74.77 \\ 

\bottomrule
\end{tabular}
\end{adjustbox}
\caption{Ablation study on various datasets and tasks.}
\label{tab:ablation_study}
\end{table*}

\subsection{Main Results}
We present the experimental results of PgM alongside model learning baselines, as shown in Table \ref{tab:main_result}.
First, we conduct experiments to single-modality scenarios using the proposed framework and compare their performance. The results show that all multimodal fusion methods outperform their single-modal counterparts, emphasizing the importance of integrating complementary information from different modalities. 

Next, we compare the performance between PgM and multimodal learning method (e.g., concatenation and add) across four multimodal tasks. Our proposed multimodal learning framework consistently surpasses all baseline multimodal learning methods across various tasks, highlighting its strong generalization and learning capabilities. For instance, in multimodal sentiment analysis, PgM outperforms baselines such as Concatenation and Add by at least 15\%-18\% points. In the multimodal emotion recognition task, PgM achieves an accuracy (ACC) of 66.69 and a weighted F1 score of 66.95, significantly outperforming baselines, particularly single-modal approaches. Also, in both cross-modal retrieval and image-text classification, PgM consistently achieves superior performance across multiple metrics and tasks, demonstrating its robustness and adaptability in multiple multimodal tasks. These findings highlight the advantages of PgM in two key aspects: (1) effective multimodal learning and (2) enhanced utilization of multimodal information, providing valuable insights for advancing multimodal learning.

\subsection{Ablation Study}
Table \ref{tab:ablation_study} presents an ablation study across multiple datasets and tasks, assessing the impact of the modal partitioner, uni-modal learner, paired-modal learner, and uni-paired modal decoder on model performance. The ablation process entails removing these modules along with their corresponding loss terms from the training objective.

Initially, we remove the modal partitioner from PgM. The uni-modal encoder, paired-modal encoder, and uni-paired modal decoder are built upon the modal partitioner. Consequently, removing the modal partitioner also eliminates the uni-modal encoder, paired-modal encoder, and uni-paired modal decoder, resulting in a significant performance drop, with accuracy (ACC) decreasing to 48.58/50.69 and weighted F1 dropping to 44.31/47.46. Next, we sequentially removed the uni-modal, paired-modal and uni-paired decoder to assess its role. Removing modal learners including uni-modal and paired-modal results in decreased metrics across all tasks. Specifically, in multimodal sentiment analysis and multimodal emotion recognition, removing the modal partitioner and modal learner resulted in reductions of at least 20\% and 10\% in all metrics, respectively. In cross-modal retrieval, the absence of these components led to a decrease in MAP and Precision@10. In the image-text classification task (UMPC Food 101), removing these components caused drops in ACC and Weighted F1 scores. These declines underscore the effectiveness of combining both modal partitioner and modal learner in multimodal learning. 
In conclusion, each module in PgM play a crucial role in significantly improving model performance across various tasks and their corresponding metrics.

\subsection{Adapting PgM for Multimodal Tasks}
We further evaluate the adaptivity of PgM on existing multimodal models, i.e., we enhance the performance of existing models with PgM. The results (enhanced performance is indicated by an underline) are given in Table \ref{tab:adaption}. 

We integrate the proposed PgM framework into well-established models (e.g., Self-MM, MMIM, UniMSE, and UniMEEC) in the MSA and MERC fields, with their enhanced versions denoted by the superscript $\dagger$. Specifically, we integrate our proposed PgM into their framework by replacing their multimodal learning modules. For MSA task, $\text{Self-MM}^{\dagger}$ demonstrates a substantial improvement, achieving 85.54/86.75 ACC-2 and 85.18/86.36 F1 score, outperforming original Self-MM's results, highlighting PgM's ability to enhance multimodal learning. 
$\text{MMIM}^{\dagger}$ also demonstrates improvement, achieving ACC-2 scores of 85.36/86.63 and F1 scores of 85.27/86.15, with approximately 0.5–1\% gains across all metrics. However, it benefits less from the proposed multimodal learning framework compared to other models. For the MERC task, $\text{UniMSE}^{\dagger}$ shows the most significant improvement, reaching 66.48 ACC and 67.06 Weighted F1, with an approximate 1–2\% boost across all metrics. Similarly, $\text{UniMEEC}^{\dagger}$ achieves the highest gains, with 67.92 ACC and 68.76 Weighted F1. These results confirm PgM’s effectiveness in enhancing already strong models and further highlight its adaptability.

\begin{table}[t]
\centering
\begin{adjustbox}{max width=0.9\linewidth}
\begin{tabular}{lccc}
\toprule
                      & \#\bf Model                & \bf ACC-2(ACC) \%        & \bf F1(Weighted F1)\%           \\ \toprule
\multirow{4}{*}{MSA}  & Self-MM                    & 84.00/85.98       & 84.42/85.95       \\
                      & $\text{Self-MM}^{\dagger}$ & \uline{85.54/86.75}\textcolor{red}{\(\uparrow\)} & \uline{85.18/86.36}\textcolor{red}{\(\uparrow\)} \\
                      & MMIM                       & 84.14/86.06       & 84.00/85.98       \\
                      & $\text{MMIM}^{\dagger}$    & \uline{85.36/86.63}\textcolor{red}{\(\uparrow\)} & \uline{85.27/86.15}\textcolor{red}{\(\uparrow\)} \\
\midrule
\multirow{4}{*}{MERC} & UniMSE                     & 65.09      & 65.51       \\
                      & $\text{UniMSE}^{\dagger}$  & \uline{66.48}\textcolor{red}{\(\uparrow\)}  & \uline{67.06}\textcolor{red}{\(\uparrow\)} \\ 
                      & UniMEEC                     &   67.36     & 68.09       \\
                      & $\text{UniMEEC}^{\dagger}$  & \uline{67.92}\textcolor{red}{\(\uparrow\)} & \uline{68.76}\textcolor{red}{\(\uparrow\)}\\
\bottomrule
\end{tabular}
\end{adjustbox}
\caption{Adapt PgM to the prior models for multimodal sentiment analysis on MOSI dataset and multimodal emotion recognition in Conversation on MELD dataset.}
\label{tab:adaption}
\end{table}

\subsection{Visualization}
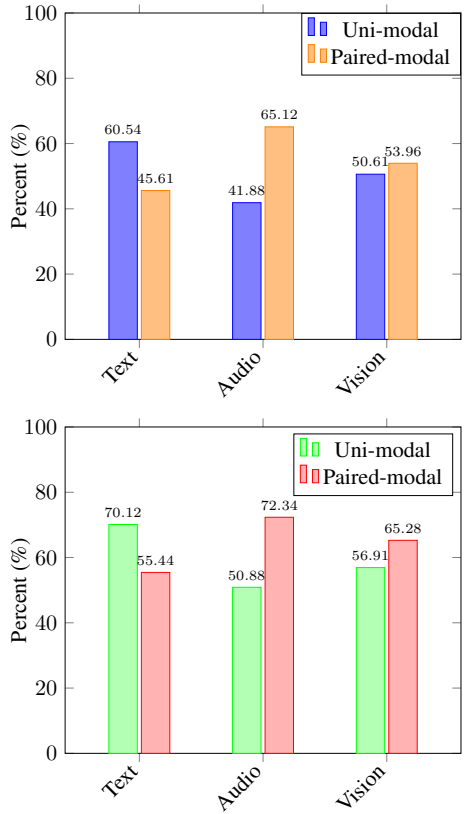
\begin{figure}[t!]
    \centering
    \begin{tikzpicture}[scale=0.76]
        \begin{axis}[
            ybar,
            bar width=14pt,
            symbolic x coords={Text, Audio, Vision},
            xtick=data,
            ymin=0, ymax=100,
            ylabel={Percent (\%)},
            ylabel style={yshift=-1.2em},
            enlarge x limits=0.3,
            xticklabel style={rotate=45, anchor=east}, % Rotate x-axis labels
            legend style={at={(1,1)}, anchor=north east},
            nodes near coords,
            every node near coord/.append style={font=\scriptsize}
        ]
        \addplot[
        fill=blue!50, 
        draw=blue   
        ] coordinates {(Text, 60.54) (Audio, 41.88) (Vision, 50.61)};
        \addplot[
        fill=orange!50, 
        draw=orange
        ] coordinates {(Text, 45.61) (Audio, 65.12) (Vision, 53.96)};
        \legend{Uni-modal, Paired-modal}
        \end{axis}
    \end{tikzpicture}
    \begin{tikzpicture}[scale=0.76]
            \begin{axis}[
                ybar,
                bar width=14pt,
                symbolic x coords={Text, Audio, Vision},
                xtick=data,
                ymin=0, ymax=100,
                ylabel={Percent (\%)},
                ylabel style={yshift=-1.2em},
                enlarge x limits=0.3,
                xticklabel style={rotate=45, anchor=east}, % Rotate x-axis labels
                nodes near coords,
                every node near coord/.append style={font=\scriptsize}
            ]
            \addplot[
            fill=green!30, 
            draw=green   
            ] coordinates {(Text, 70.12) (Audio, 50.88) (Vision, 56.91)};
            \addplot[
            fill=red!30, 
            draw=red
            ] coordinates {(Text, 55.44) (Audio, 72.34) (Vision, 65.28)};
            \legend{Uni-modal, Paired-modal}
            \end{axis}
        \end{tikzpicture}
    \caption{Distribution of uni-modal and paired-modal features across text, audio and vision modalities for multimodal sentiment analysis (top) and multimodal emotion recognition in conversation (bottom).}
    \label{fig:distribution_feature}
\end{figure}

Furthermore, we visualize the distribution of uni-modal and paired-modal features for each individual modality after applying the modal partitioner, derived from both MSA and MERC tasks, as shown in Figure \ref{fig:distribution_feature}. 

In the text modality, 60.54\% of features are uni-modal, and 45.61\% are paired-modal, indicating that uni-modal features independently provide more information for the MSA task. In the audio modality, 41.88\% are uni-modal and 65.12\% are paired-modal, suggesting that its paired-modal features play a larger role in the MSA task. In the vision modality, uni-modal and paired-modal features are nearly evenly distributed (50.61\% vs 53.96\%), indicating that both uni-modal and paired-modal features contribute to multimodal sentiment analysis. Similarly, we analyze the distribution of uni-modal and paired-modal features across modalities in the MERC task, which differs significantly from the MSA task. For example, in MERC, uni-modal features in the text modality remain the dominant source of information, while paired-modal features in the audio and vision modalities contribute more compared to their respective uni-modal features. These findings highlight that modal partitioner in PgM adjusts the distribution of uni-modal and paired-modal features across different multimodal tasks.

%%%%%%%%%%%%%%%%%%%%%%%%%%%%%%%%%%%%%%%%%%%%%%%%%%%%%%%%%%%%%%%%%%%%%%%
\section{Conclusion}
This paper presents PgM\includegraphics[width=1.25em]{figures/cut_2.png}, a partitioner-guided modal learning framework consisting of the modal partitioner, modal learner, and uni-paired modal decoder. The modal partitioner divides the modal representation into uni-modal and paired-modal features, the modal learner enhances their learning through dedicated uni-modal and paired-modal components, and the uni-paired decoder reconstructs the modal representation. Extensive experiments on four multimodal tasks—spanning vision, language, and audio modalities—demonstrate PgM’s versatility and effectiveness. Additionally, we visualize the contributions of uni-modal and paired-modal features to multimodal tasks, offering valuable insights into their respective roles. We believe this work presents a new experimental setting that can provide a new and different perspective to multimodal learning communities.
\section*{Limitations}
This study identifies three main limitations of PgM. First, PgM operates on each modality, incorporating multiple learning modules, which increases the number of training parameters. A detailed analysis of the parameter size of PgM is provided in Appendix \ref{sec:model_size}. Second, PgM is designed to enhance multimodal representation learning, with its training process separate from that of downstream multimodal tasks. As a result, integrating PgM with downstream task training requires two stages: pretraining PgM and jointly training both the downstream task and PgM. However, the separate training process also offers adaptive flexibility for various multimodal tasks. Third, PgM focuses on three primary modalities—text, vision, and audio—while overlooking other modalities such as sensor and biometric data. Although PgM achieves strong performance gains, the downstream fusion mechanism remains relatively simple (i.e., concatenation followed by a feed-forward network), which may limit the full potential of the partitioned features. In future work, we plan to explore more structured approaches to better preserve and utilize the separation between uni-modal and paired-modal representations.

\section*{Ethics Statement}
The data used in this study consists entirely of open-source datasets for research purposes. While our approach improves performance across multiple multimodal downstream tasks, it remains primarily suited for multimodal understanding tasks (e.g., classification and regression). However, even on relatively straightforward tasks such as 6-class emotion recognition (MELD), the proposed method achieves only 66\% accuracy, indicating substantial room for improvement.
\bibliography{acl_latex}

\begin{thebibliography}{30}
\expandafter\ifx\csname natexlab\endcsname\relax\def\natexlab#1{#1}\fi

\bibitem[{Arandjelovic and Zisserman(2017)}]{DBLP:conf/iccv/ArandjelovicZ17}
Relja Arandjelovic and Andrew Zisserman. 2017.
\newblock \href {https://doi.org/10.1109/ICCV.2017.73} {Look, listen and learn}.
\newblock In \emph{{IEEE} International Conference on Computer Vision, {ICCV} 2017, Venice, Italy, October 22-29, 2017}, pages 609--617. {IEEE} Computer Society.

\bibitem[{Du et~al.(2023)Du, Teng, Li, Liu, Yuan, Wang, Yuan, and Zhao}]{DBLP:conf/icml/DuTLLYWYZ23}
Chenzhuang Du, Jiaye Teng, Tingle Li, Yichen Liu, Tianyuan Yuan, Yue Wang, Yang Yuan, and Hang Zhao. 2023.
\newblock \href {https://proceedings.mlr.press/v202/du23e.html} {On uni-modal feature learning in supervised multi-modal learning}.
\newblock In \emph{International Conference on Machine Learning, {ICML} 2023, 23-29 July 2023, Honolulu, Hawaii, {USA}}, volume 202 of \emph{Proceedings of Machine Learning Research}, pages 8632--8656. {PMLR}.

\bibitem[{Goyal et~al.(2017)Goyal, Khot, Summers{-}Stay, Batra, and Parikh}]{DBLP:conf/cvpr/GoyalKSBP17}
Yash Goyal, Tejas Khot, Douglas Summers{-}Stay, Dhruv Batra, and Devi Parikh. 2017.
\newblock \href {https://doi.org/10.1109/CVPR.2017.670} {Making the {V} in {VQA} matter: Elevating the role of image understanding in visual question answering}.
\newblock In \emph{2017 {IEEE} Conference on Computer Vision and Pattern Recognition, {CVPR} 2017, Honolulu, HI, USA, July 21-26, 2017}, pages 6325--6334. {IEEE} Computer Society.

\bibitem[{Grover et~al.(2023)Grover, Chharia, Upadhyay, and Longo}]{10023506}
Nitin Grover, Aviral Chharia, Rahul Upadhyay, and Luca Longo. 2023.
\newblock \href {https://doi.org/10.1109/TNSRE.2023.3237375} {Schizo-net: A novel schizophrenia diagnosis framework using late fusion multimodal deep learning on electroencephalogram-based brain connectivity indices}.
\newblock \emph{IEEE Transactions on Neural Systems and Rehabilitation Engineering}, 31:464--473.

\bibitem[{Han et~al.()Han, Chen, and Poria}]{DBLP:conf/emnlp/HanCP21}
Wei Han, Hui Chen, and Soujanya Poria.
\newblock Improving multimodal fusion with hierarchical mutual information maximization for multimodal sentiment analysis.
\newblock In \emph{Proceedings of the 2021 Conference on Empirical Methods in Natural Language Processing, {EMNLP} 2021, Virtual Event / Punta Cana, Dominican Republic, 7-11 November, 2021}, pages 9180--9192.

\bibitem[{Hu et~al.(2022)Hu, Lin, Zhao, Lu, Wu, and Li}]{hu2022unimse}
Guimin Hu, Ting-En Lin, Yi~Zhao, Guangming Lu, Yuchuan Wu, and Yongbin Li. 2022.
\newblock Unimse: Towards unified multimodal sentiment analysis and emotion recognition.
\newblock \emph{arXiv preprint arXiv:2211.11256}.

\bibitem[{Hu et~al.(2024)Hu, Zhu, Hershcovich, Hu, Seifi, and Xie}]{hu2024unimeec}
Guimin Hu, Zhihong Zhu, Daniel Hershcovich, Lijie Hu, Hasti Seifi, and Jiayuan Xie. 2024.
\newblock Unimeec: Towards unified multimodal emotion recognition and emotion cause.
\newblock \emph{arXiv preprint arXiv:2404.00403}.

\bibitem[{Huang et~al.(2021)Huang, Du, Xue, Chen, Zhao, and Huang}]{DBLP:conf/nips/HuangDXCZH21}
Yu~Huang, Chenzhuang Du, Zihui Xue, Xuanyao Chen, Hang Zhao, and Longbo Huang. 2021.
\newblock \href {https://proceedings.neurips.cc/paper/2021/hash/5aa3405a3f865c10f420a4a7b55cbff3-Abstract.html} {What makes multi-modal learning better than single (provably)}.
\newblock In \emph{Advances in Neural Information Processing Systems 34: Annual Conference on Neural Information Processing Systems 2021, NeurIPS 2021, December 6-14, 2021, virtual}, pages 10944--10956.

\bibitem[{Huang et~al.(2022)Huang, Lin, Zhou, Yang, and Huang}]{DBLP:conf/icml/HuangLZYH22}
Yu~Huang, Junyang Lin, Chang Zhou, Hongxia Yang, and Longbo Huang. 2022.
\newblock \href {https://proceedings.mlr.press/v162/huang22e.html} {Modality competition: What makes joint training of multi-modal network fail in deep learning? (provably)}.
\newblock In \emph{International Conference on Machine Learning, {ICML} 2022, 17-23 July 2022, Baltimore, Maryland, {USA}}, volume 162 of \emph{Proceedings of Machine Learning Research}, pages 9226--9259. {PMLR}.

\bibitem[{Li et~al.(2023)Li, Li, Savarese, and Hoi}]{li2023blip}
Junnan Li, Dongxu Li, Silvio Savarese, and Steven Hoi. 2023.
\newblock Blip-2: Bootstrapping language-image pre-training with frozen image encoders and large language models.
\newblock In \emph{International conference on machine learning}, pages 19730--19742. PMLR.

\bibitem[{Li et~al.(2022)Li, Li, Xiong, and Hoi}]{li2022blip}
Junnan Li, Dongxu Li, Caiming Xiong, and Steven Hoi. 2022.
\newblock Blip: Bootstrapping language-image pre-training for unified vision-language understanding and generation.
\newblock In \emph{International conference on machine learning}, pages 12888--12900. PMLR.

\bibitem[{Li et~al.(2021)Li, Selvaraju, Gotmare, Joty, Xiong, and Hoi}]{li2021align}
Junnan Li, Ramprasaath Selvaraju, Akhilesh Gotmare, Shafiq Joty, Caiming Xiong, and Steven Chu~Hong Hoi. 2021.
\newblock Align before fuse: Vision and language representation learning with momentum distillation.
\newblock \emph{Advances in neural information processing systems}, 34:9694--9705.

\bibitem[{Li et~al.(2024)Li, Zhang, Li, Peng, Tang, Zhou, Zhang, Hu, Yuan, S{\o}gaard et~al.}]{li2024foodieqa}
Wenyan Li, Crystina Zhang, Jiaang Li, Qiwei Peng, Raphael Tang, Li~Zhou, Weijia Zhang, Guimin Hu, Yifei Yuan, Anders S{\o}gaard, et~al. 2024.
\newblock Foodieqa: A multimodal dataset for fine-grained understanding of chinese food culture.
\newblock In \emph{Proceedings of the 2024 Conference on Empirical Methods in Natural Language Processing}, pages 19077--19095.

\bibitem[{Lian et~al.(2024)Lian, Sun, Sun, Wen, Zhang, Chen, Gu, Zhao, Ma, Chen et~al.}]{lian2024mer}
Zheng Lian, Haiyang Sun, Licai Sun, Zhuofan Wen, Siyuan Zhang, Shun Chen, Hao Gu, Jinming Zhao, Ziyang Ma, Xie Chen, et~al. 2024.
\newblock Mer 2024: Semi-supervised learning, noise robustness, and open-vocabulary multimodal emotion recognition.
\newblock In \emph{Proceedings of the 2nd International Workshop on Multimodal and Responsible Affective Computing}, pages 41--48.

\bibitem[{Liang et~al.(2022)Liang, Lyu, Chhablani, Jain, Deng, Wang, Morency, and Salakhutdinov}]{DBLP:journals/corr/abs-2207-00056}
Paul~Pu Liang, Yiwei Lyu, Gunjan Chhablani, Nihal Jain, Zihao Deng, Xingbo Wang, Louis{-}Philippe Morency, and Ruslan Salakhutdinov. 2022.
\newblock \href {https://doi.org/10.48550/ARXIV.2207.00056} {Multiviz: An analysis benchmark for visualizing and understanding multimodal models}.
\newblock \emph{CoRR}, abs/2207.00056.

\bibitem[{Mao et~al.(2022)Mao, Yuan, Xu, Yu, Liu, and Gao}]{mao2022m}
Huisheng Mao, Ziqi Yuan, Hua Xu, Wenmeng Yu, Yihe Liu, and Kai Gao. 2022.
\newblock M-sena: An integrated platform for multimodal sentiment analysis.
\newblock \emph{arXiv preprint arXiv:2203.12441}.

\bibitem[{P{\'e}rez-R{\'u}a et~al.(2019)P{\'e}rez-R{\'u}a, Vielzeuf, Pateux, Baccouche, and Jurie}]{perez2019mfas}
Juan-Manuel P{\'e}rez-R{\'u}a, Valentin Vielzeuf, St{\'e}phane Pateux, Moez Baccouche, and Fr{\'e}d{\'e}ric Jurie. 2019.
\newblock Mfas: Multimodal fusion architecture search.
\newblock In \emph{Proceedings of the IEEE/CVF Conference on Computer Vision and Pattern Recognition}, pages 6966--6975.

\bibitem[{Poria et~al.(2019)Poria, Hazarika, Majumder, Naik, Cambria, and Mihalcea}]{DBLP:conf/acl/PoriaHMNCM19}
Soujanya Poria, Devamanyu Hazarika, Navonil Majumder, Gautam Naik, Erik Cambria, and Rada Mihalcea. 2019.
\newblock \href {https://doi.org/10.18653/v1/p19-1050} {{MELD:} {A} multimodal multi-party dataset for emotion recognition in conversations}.
\newblock In \emph{Proceedings of the 57th Conference of the Association for Computational Linguistics, {ACL} 2019, Florence, Italy, July 28- August 2, 2019, Volume 1: Long Papers}, pages 527--536. Association for Computational Linguistics.

\bibitem[{Rasiwasia et~al.(2010)Rasiwasia, Costa~Pereira, Coviello, Doyle, Lanckriet, Levy, and Vasconcelos}]{rasiwasia2010new}
Nikhil Rasiwasia, Jose Costa~Pereira, Emanuele Coviello, Gabriel Doyle, Gert~RG Lanckriet, Roger Levy, and Nuno Vasconcelos. 2010.
\newblock A new approach to cross-modal multimedia retrieval.
\newblock In \emph{Proceedings of the 18th ACM international conference on Multimedia}, pages 251--260.

\bibitem[{Singh et~al.(2022)Singh, Hu, Goswami, Couairon, Galuba, Rohrbach, and Kiela}]{singh2022flava}
Amanpreet Singh, Ronghang Hu, Vedanuj Goswami, Guillaume Couairon, Wojciech Galuba, Marcus Rohrbach, and Douwe Kiela. 2022.
\newblock Flava: A foundational language and vision alignment model.
\newblock In \emph{Proceedings of the IEEE/CVF Conference on Computer Vision and Pattern Recognition}, pages 15638--15650.

\bibitem[{Tsai et~al.(2019)Tsai, Bai, Liang, Kolter, Morency, and Salakhutdinov}]{DBLP:conf/acl/TsaiBLKMS19}
Yao{-}Hung~Hubert Tsai, Shaojie Bai, Paul~Pu Liang, J.~Zico Kolter, Louis{-}Philippe Morency, and Ruslan Salakhutdinov. 2019.
\newblock \href {https://doi.org/10.18653/v1/p19-1656} {Multimodal transformer for unaligned multimodal language sequences}.
\newblock In \emph{Proceedings of the 57th Conference of the Association for Computational Linguistics, {ACL} 2019, Florence, Italy, July 28- August 2, 2019, Volume 1: Long Papers}, pages 6558--6569. Association for Computational Linguistics.

\bibitem[{Vaswani(2017)}]{vaswani2017attention}
A~Vaswani. 2017.
\newblock Attention is all you need.
\newblock \emph{Advances in Neural Information Processing Systems}.

\bibitem[{Wang et~al.(2020)Wang, Tran, and Feiszli}]{DBLP:conf/cvpr/WangTF20}
Weiyao Wang, Du~Tran, and Matt Feiszli. 2020.
\newblock \href {https://doi.org/10.1109/CVPR42600.2020.01271} {What makes training multi-modal classification networks hard?}
\newblock In \emph{2020 {IEEE/CVF} Conference on Computer Vision and Pattern Recognition, {CVPR} 2020, Seattle, WA, USA, June 13-19, 2020}, pages 12692--12702. Computer Vision Foundation / {IEEE}.

\bibitem[{Wang et~al.(2015)Wang, Kumar, Thome, Cord, and Precioso}]{wang2015recipe}
Xin Wang, Devinder Kumar, Nicolas Thome, Matthieu Cord, and Frederic Precioso. 2015.
\newblock Recipe recognition with large multimodal food dataset.
\newblock In \emph{2015 IEEE International Conference on Multimedia \& Expo Workshops (ICMEW)}, pages 1--6. IEEE.

\bibitem[{Xie et~al.(2025)Xie, Cheng, Zhang, Cai, Hu, Xie, and Li}]{xie2025explicitly}
Jiayuan Xie, Mengqiu Cheng, Xinting Zhang, Yi~Cai, Guimin Hu, Mengying Xie, and Qing Li. 2025.
\newblock Explicitly guided difficulty-controllable visual question generation.
\newblock In \emph{Proceedings of the AAAI Conference on Artificial Intelligence}, volume~39, pages 25552--25560.

\bibitem[{Yu et~al.(2021)Yu, Xu, Yuan, and Wu}]{DBLP:conf/aaai/YuXYW21}
Wenmeng Yu, Hua Xu, Ziqi Yuan, and Jiele Wu. 2021.
\newblock Learning modality-specific representations with self-supervised multi-task learning for multimodal sentiment analysis.
\newblock In \emph{Thirty-Fifth {AAAI} Conference on Artificial Intelligence, {AAAI} 2021, Thirty-Third Conference on Innovative Applications of Artificial Intelligence, {IAAI} 2021, The Eleventh Symposium on Educational Advances in Artificial Intelligence, {EAAI} 2021, Virtual Event, February 2-9, 2021}, pages 10790--10797. {AAAI} Press.

\bibitem[{Zadeh et~al.(2016)Zadeh, Zellers, Pincus, and Morency}]{DBLP:journals/expert/ZadehZPM16}
Amir Zadeh, Rowan Zellers, Eli Pincus, and Louis{-}Philippe Morency. 2016.
\newblock \href {https://doi.org/10.1109/MIS.2016.94} {Multimodal sentiment intensity analysis in videos: Facial gestures and verbal messages}.
\newblock \emph{{IEEE} Intell. Syst.}, 31(6):82--88.

\bibitem[{Zhang et~al.(2022)Zhang, Chen, Shen, and Wang}]{DBLP:conf/aaai/ZhangCSW22}
Yi~Zhang, Mingyuan Chen, Jundong Shen, and Chongjun Wang. 2022.
\newblock \href {https://doi.org/10.1609/AAAI.V36I8.20895} {Tailor versatile multi-modal learning for multi-label emotion recognition}.
\newblock In \emph{Thirty-Sixth {AAAI} Conference on Artificial Intelligence, {AAAI} 2022, Thirty-Fourth Conference on Innovative Applications of Artificial Intelligence, {IAAI} 2022, The Twelveth Symposium on Educational Advances in Artificial Intelligence, {EAAI} 2022 Virtual Event, February 22 - March 1, 2022}, pages 9100--9108. {AAAI} Press.

\bibitem[{Zhu et~al.(2024{\natexlab{a}})Zhu, Cheng, Hu, Li, Huang, and Zou}]{zhu2024towards}
Zhihong Zhu, Xuxin Cheng, Guimin Hu, Yaowei Li, Zhiqi Huang, and Yuexian Zou. 2024{\natexlab{a}}.
\newblock Towards multi-modal sarcasm detection via disentangled multi-grained multi-modal distilling.
\newblock In \emph{Proceedings of the 2024 Joint International Conference on Computational Linguistics, Language Resources and Evaluation (LREC-COLING 2024)}, pages 16581--16591.

\bibitem[{Zhu et~al.(2024{\natexlab{b}})Zhu, Zhuang, Zhang, Xu, Hu, Wu, and Zheng}]{zhu2024tfcd}
Zhihong Zhu, Xianwei Zhuang, Yunyan Zhang, Derong Xu, Guimin Hu, Xian Wu, and Yefeng Zheng. 2024{\natexlab{b}}.
\newblock Tfcd: Towards multi-modal sarcasm detection via training-free counterfactual debiasing.
\newblock In \emph{Proc. of IJCAI}.

\end{thebibliography}

\newpage
\appendix

% \section{Appendix}
\section{Detailed Architecture of PgM}
\label{sec:architecture_pgm}
In this section, we will introduce the architecture of PgM. We first provide the components of PgM and then show the specific architecture.

\paragraph{Modal Encoder:} We adopt T5 as text encoder, AST as audio encoder and ViT as image encoder. Each modal encoder take raw modal information as input and outputs modal representation as shape (B,S,D), where B denotes batch size, S denotes sequence length, and D denotes the dimension of modal representation. 

\paragraph{Modal Partitioner:} The modal partitioner takes each individual modal representation as input and outputs uni-modal and paired-modal gates with a shape of (B, S, D). Furthermore, we represent uni-modal and paired-modal gates as a vector, where each element belongs to $\{0,1\}$. These gates are then used to derive padding masks for the uni-modal and paired-modal learners, respectively. 

\paragraph{Uni-Modal Learner:} The uni-modal learner learns uni-modal features by masking paired-modal features, which are indicated by $-\infty$ in the padding mask. The padding mask has a shape of (B, S, D) corresponding to the uni-modal learner.

\paragraph{Paired-Modal Learner:} The paired-modal learner learns paired-modal features by masking uni-modal features, which are indicated by $-\infty$ in the padding mask of the paired-modal learner. The padding mask has a shape of (B, S, D) corresponding to the paired-modal representation.

\paragraph{Uni-Paired Decoder:} The uni-paired modal decoder takes the concatenation of the uni-modal and paired-modal representations as input, then passes it through the subsequent FFN layer to reconstruct the original modal representation.

\paragraph{Padding Mask} We derive padding masks for uni-modal and paired-modal learners from $\hat{\mathbf{g}}^{(i)}_{u}$ and $\hat{\mathbf{g}}^{(i)}_{p}$, respectively. For instance, given $\hat{\mathbf{g}}^{(i)}_{u} \in \mathcal{R}^{1\times D}$ with $\hat{\mathbf{g}}^{(i)}_{u} = [1,1,1,1,0,0]$, the first four positions indicate the features the uni-modal learner should focus on. Accordingly, its padding mask $\mathbf{M}^{u} \in \mathcal{R}^{S\times D}$ is generated through a broadcasting operation on $\hat{\mathbf{g}}^{(i)}_{u}$, where $S$ and $D$ denote the sequence length and representation dimension, respectively:
\begin{align*}
\mathbf{M}^{u} =\begin{bmatrix}
0 & 0 & 0 & 0 &  -\infty & -\infty \\
0 &  0 & 0 & 0 &  -\infty & -\infty \\
0 & 0 & 0 & 0 &  -\infty& -\infty \\
0 & 0 & 0 & 0 &  -\infty& -\infty \\
\end{bmatrix}
\end{align*}
Similarly, we also can obtain padding mask $\mathbf{M}^{p} \in \mathcal{R}^{S\times D}$ for paired-modal learner.
\section{Downstream Task Training}
\label{sec:downstream_training}
\begin{figure}[h]
\centering
\includegraphics[width=0.95\linewidth]{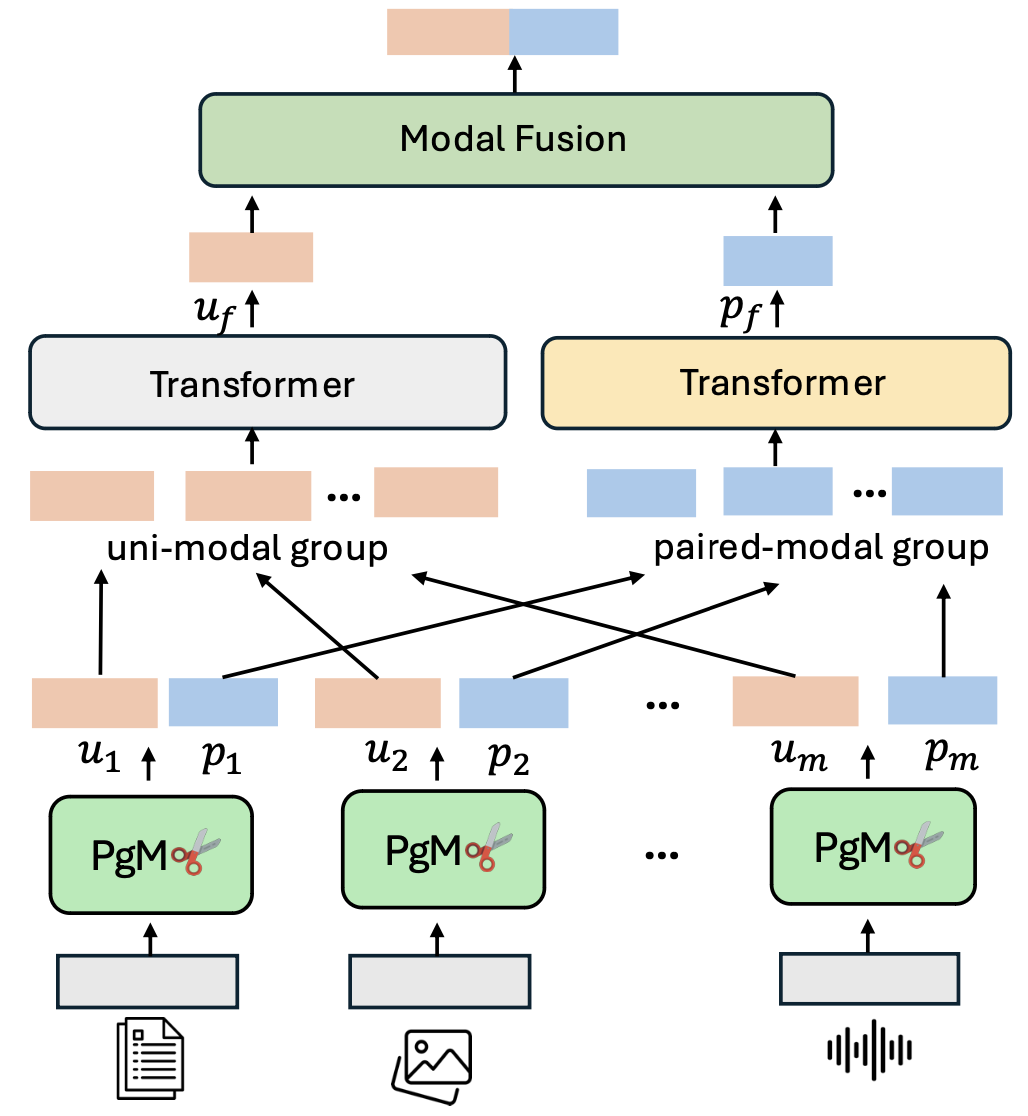}
\caption{Overview of training framework with multimodal downstream task.}
\label{fig:training}
\end{figure}
\subsection{Detail Architecture}
The architecture for jointly training PgM and the downstream task is shown in Figure \ref{fig:training}. During the training phase, the first $N^{1st}$ epochs are dedicated to PgM training, while $\text{N}^{2st}$ epochs are used for joint training the downstream task with PgM. Specifically, PgM enables us to extract uni-modal and paired-modal features for each modality. We then categorize these features into two groups: uni-modal and paired-modal features across all modalities. We independently concatenate all uni-modal and paired-modal features from different modalities within a single sample and pass the combined representations through two FFN layers respectively. Next, we feed the concatenated uni-modal and paired-modal representations into two separate Transformers, where each serves as the query, key, and value. This process generates the final uni-modal and paired-modal representations, each incorporating features from all modalities. Given a sample with modalities $\{m_{1}, m_{2}\}$, the following modules in downstream task training are as follows:
\begin{align*}
   &\hat{\mathbf{p}}=\text{FFN}([\mathbf{p}_{m_{1}}, \mathbf{p}_{m_{1}}])\\
   &\hat{\mathbf{u}}=\text{FFN}([\mathbf{u}_{m_{1}}, \mathbf{u}_{m_{1}}])\\
   &\mathbf{p}_{f} = \text{Transformer}(\hat{\mathbf{p}},\hat{\mathbf{p}},\hat{\mathbf{p}}))\\
    &\mathbf{u}_{f} = \text{Transformer}(\hat{\mathbf{u}},\hat{\mathbf{u}},\hat{\mathbf{u}}))
\end{align*}
Here, $\{\mathbf{p}_{m_{1}}, \mathbf{p}_{m_{2}}\}$ represent the paired-modal features from modalities $\{m_{1}, m_{2}\}$, while $\{\mathbf{u}_{m_{1}}, \mathbf{u}_{m_{2}}\}$ represent the uni-modal features from modalities $\{m_{1}, m_{2}\}$. $\mathbf{u}_{f}$ and $\mathbf{p}_{f}$ denote the uni-modal and paired-modal representations, respectively, formed by combining multiple paired-modal features from different modalities.

Next, we concatenate the paired-modal and uni-modal representations, each integrating multiple modalities within a sample, and pass the combined representations through two FFN layers:
\begin{align}
   &\hat{\mathbf{I}}=\text{FFN}([\mathbf{u}_{f}, \mathbf{p}_{f}])\\
   &\hat{\mathbf{y}} = \text{Prediction}(\hat{\mathbf{I}})\label{eq:(18)}
\end{align}
where $\hat{\mathbf{y}}$ denotes prediction results on categories. $\hat{\mathbf{I}}$ denotes the final multimodal fusion representation, where we use concatenation as the modal fusion method. The goal of the modal fusion module is to combine the uni-modal and paired-modal representations into a single vector. This can be achieved through fusion methods such as concatenation, addition, or more complex structures. In our work, we use concatenation as the fusion method. The {\bf Prediction} in Equation (\ref{eq:(18)}) represents the classifier for classification tasks and the regression layers for regression tasks, respectively.

\section{Additional Experiments}

\subsection{Model Size}
\label{sec:model_size}
To further improve the understanding of PgM, we quantify the training parameter sizes for each module. The sizes of trainable parameters for each module are reported in Table \ref{tab:training_parameters}. PgM comprises five modules: (1) modal encoder, (2) modal partitioner, (3) uni-modal learner, (4) paired-modal learner, and (5) uni-paired modal decoder. During the training phase, the parameters of the modal encoders remain fixed, while only those in the modal partitioner, uni-modal learner, paired-modal learner, and uni-paired decoder are updated based on the training loss. The modal encoder has no trainable parameters since we utilize a frozen pre-trained model. The modal partitioner requires only a small number of trainable parameters, highlighting its lightweight nature in feature partitioning. In contrast, the uni-modal learner, paired-modal learner, and uni-paired decoder adopt a Transformer-based architecture, each containing approximately 7.1M trainable parameters.
\begin{table}[!t]
\centering
\begin{tabular}{lc}
\toprule
Modules & Trainable Parameter  \\
\toprule
Modal encoder & 0M  \\
Modal partitioner & 1.1M  \\
Uni-modal learner &  7.08M \\
Paired-modal learner & 7.08M  \\
Uni-Paired decoder & 7.1M  \\
PgM & 22.36M \\
\bottomrule
\end{tabular}
\caption{The number of trainable parameters in each module of PgM.}
\label{tab:training_parameters}
\end{table}
\begin{figure}[t]
\centering
\includegraphics[width=0.95\linewidth]{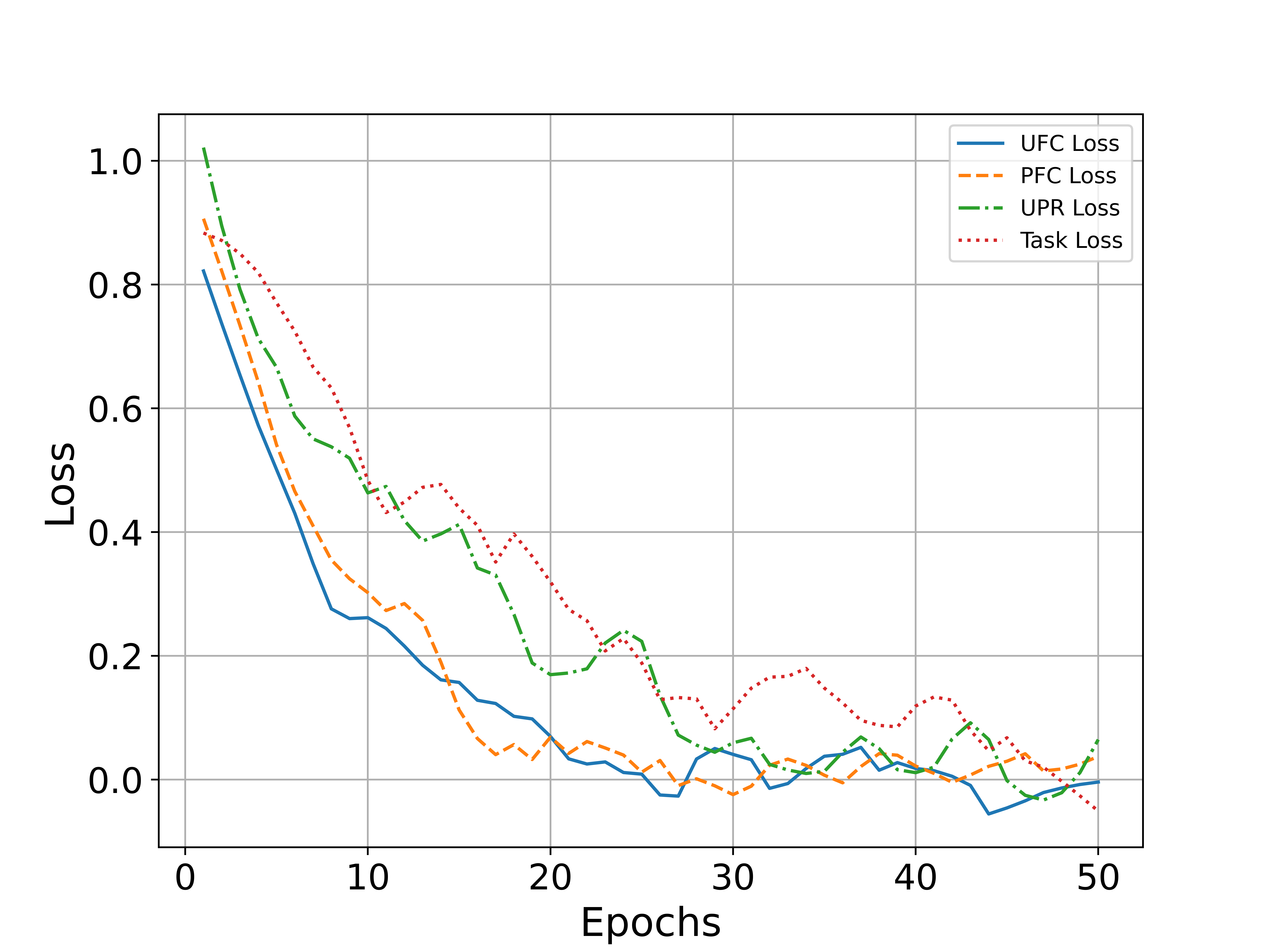}
\caption{Loss variation curves during the second stage, i.e., the joint training process of the downstream task and PgM for multimodal sentiment analysis.}
\label{fig:pgm_loss}
\end{figure}
\subsection{Training objectives of PgM}
\label{sec:loss_curve}
Figure \ref{fig:pgm_loss} illustrates the loss variation curves for the joint training process of the downstream task with PgM. 

In the early training stages (first 10 epochs), the losses decrease rapidly, suggesting that the model quickly captures essential features. UFC Loss (solid blue line) and PFC Loss (dashed orange line) exhibit a steep decline and stabilize after approximately 20 epochs, implying that these components learn efficiently.
UPR Loss (dash-dot green line) and Task Loss (dotted red line) decrease at a slower rate and exhibit fluctuations even after 20 epochs, suggesting that these losses are influenced by more complex modality interactions. By around epoch 30, all loss terms stabilize and approach zero, indicating that the model has largely converged. However, between epochs 30 and 50, Task Loss and UPR Loss still exhibit fluctuations. These loss curves reveal that UFC, PFC, UPR, and Task Loss decrease at different rates, highlighting PgM's varying learning dynamics across uni-modal feature, paired-modal feature, and downstream tasks. Furthermore, these results indicate that PgM mitigates modality laziness during training.

\end{document}